\documentclass{article} 
\usepackage{collas2024_conference,times}
\usepackage{easyReview}
\usepackage{soul}
\usepackage{color}

\usepackage{amsmath,amsfonts,bm}

\def\eqref#1{equation~\ref{#1}}

\def\1{\bm{1}}

\DeclareMathAlphabet{\mathsfit}{\encodingdefault}{\sfdefault}{m}{sl}
\SetMathAlphabet{\mathsfit}{bold}{\encodingdefault}{\sfdefault}{bx}{n}

\usepackage{hyperref}
\hypersetup{
    colorlinks=true,
    linkcolor=red,
    filecolor=magenta,
    urlcolor=blue,
    citecolor=purple,
    pdftitle={Overleaf Example},
    pdfpagemode=FullScreen,
    }
\usepackage[capitalise]{cleveref}
\usepackage{subcaption}
\usepackage{tcolorbox}
\tcbuselibrary{skins}
\usepackage{xcolor,colortbl}
\definecolor{darkgreen}{rgb}{0.0, 0.5, 0.0}
\definecolor{LigthGray}{gray}{0.8}
\usepackage{diagbox}
\usepackage{floatrow}
\usepackage{wrapfig}
\usepackage{lipsum}
\usepackage{colortbl}
\usepackage{nicematrix}
\usepackage{booktabs}
\usepackage{enumitem}

\title{Sub-goal Distillation: A Method to Improve Small Language Agents}

\author{Maryam Hashemzadeh\thanks{Corresponding author: Maryam Hashemzadeh. $\dagger$equal supervision.} \\
Mila – Quebec AI Institute, Université de Montréal\\
\texttt{maryam.hashemzadeh@mila.quebec} \\
\And
Elias Stengel-Eskin \\
UNC Chapel Hill\\
\texttt{esteng@cs.unc.edu}
\And 
Sarath Chandar$^{\dagger}$  \\
Mila – Quebec AI Institute, Polytechnique Montréal, CIFAR AI Chair \\
\texttt{sarath.chandar@mila.quebec} \\
\And 
Marc-Alexandre Côté$^{\dagger}$\\
Microsoft Research, Montréal \\
\texttt{macote@microsoft.com}
}

\collasfinalcopy 

\begin{document}

\maketitle

\begin{abstract}
While Large Language Models (LLMs) have demonstrated significant promise as agents in interactive tasks, their substantial computational requirements and restricted number of calls constrain their practical utility, especially in long-horizon interactive tasks such as decision-making or in scenarios involving continuous ongoing tasks. To address these constraints, we propose a method for transferring the performance of an LLM with billions of parameters to a much smaller language model (770M parameters).  Our approach involves constructing a hierarchical agent comprising a \emph{planning module}, which learns through Knowledge Distillation from an LLM to generate sub-goals, and an \emph{execution module}, which learns to accomplish these sub-goals using elementary actions. In detail, we leverage an LLM to annotate an oracle path with a sequence of sub-goals towards completing a goal. Subsequently, we utilize this annotated data to fine-tune both the planning and execution modules. Importantly, neither module relies on real-time access to an LLM during inference, significantly reducing the overall cost associated with LLM interactions to a \emph{fixed cost}. In ScienceWorld, a challenging and multi-task interactive text environment, our method surpasses standard imitation learning based solely on elementary actions by 16.7\% (absolute). Our analysis highlights the efficiency of our approach compared to other LLM-based methods. Our code and annotated data for distillation can be found on GitHub\footnote{\url{https://github.com/chandar-lab/SubGoal_Distillation_LLM}}.

\end{abstract}

\section{Introduction}

Recently, Large Language Models (LLMs) have found applications in various fields, including multi-task learning, decision making, answering questions, summarizing documents, translating languages, completing sentences, and serving as search assistants. They showcase a remarkable ability to make predictions based on input, enabling their use in generative AI applications to produce content based on input prompts~\citep{devlin2018bert, brown2020language, rae2021scaling, chowdhery2023palm, workshop2022bloom, patel2021mapping, han2021pre, bommasani2021opportunities}.

The promising advantage of LLMs is attributed to their training on extensive text datasets, resulting in impressive capabilities. This prior knowledge can be leveraged for action planning to solve tasks in robotics and reinforcement learning~\citep{huang2022inner, brohan2023can, liang2023code}. Recent works have utilized in-context learning with LLMs to provide actions in autonomous decision-making agents and interactive environments~\citep{mahowald2023dissociating, yao2022react, schick2023toolformer, shen2023hugginggpt, nakano2021webgpt, park2023generative, lin2023swiftsage, brohan2023can}. 

However, the extreme size of LLMs makes them computationally unaffordable for many applications. Moreover, closed-source models like ChatGPT~\citep{chatgpt} and GPT-4~\citep{Gpt-4} limit accessibility and reproducibility.
Consequently, there is an increasing demand to find approaches that are less computationally intensive while still capitalizing on the knowledge embedded in LLMs. One prevalent technique is the use of Knowledge Distillation (KD)~\citep{bucilua2006mode, hinton2015distilling}, wherein a smaller model is trained with guidance from a larger model. Through this approach, we can leverage the knowledge in an LLM to train a more compact model with a reduced number of parameters.

\begin{wrapfigure}[40]{R}{0.46\textwidth}
\centering
\includegraphics[clip, trim=0.2cm 6.0cm 0.0cm 0.0cm, width=0.9\textwidth]{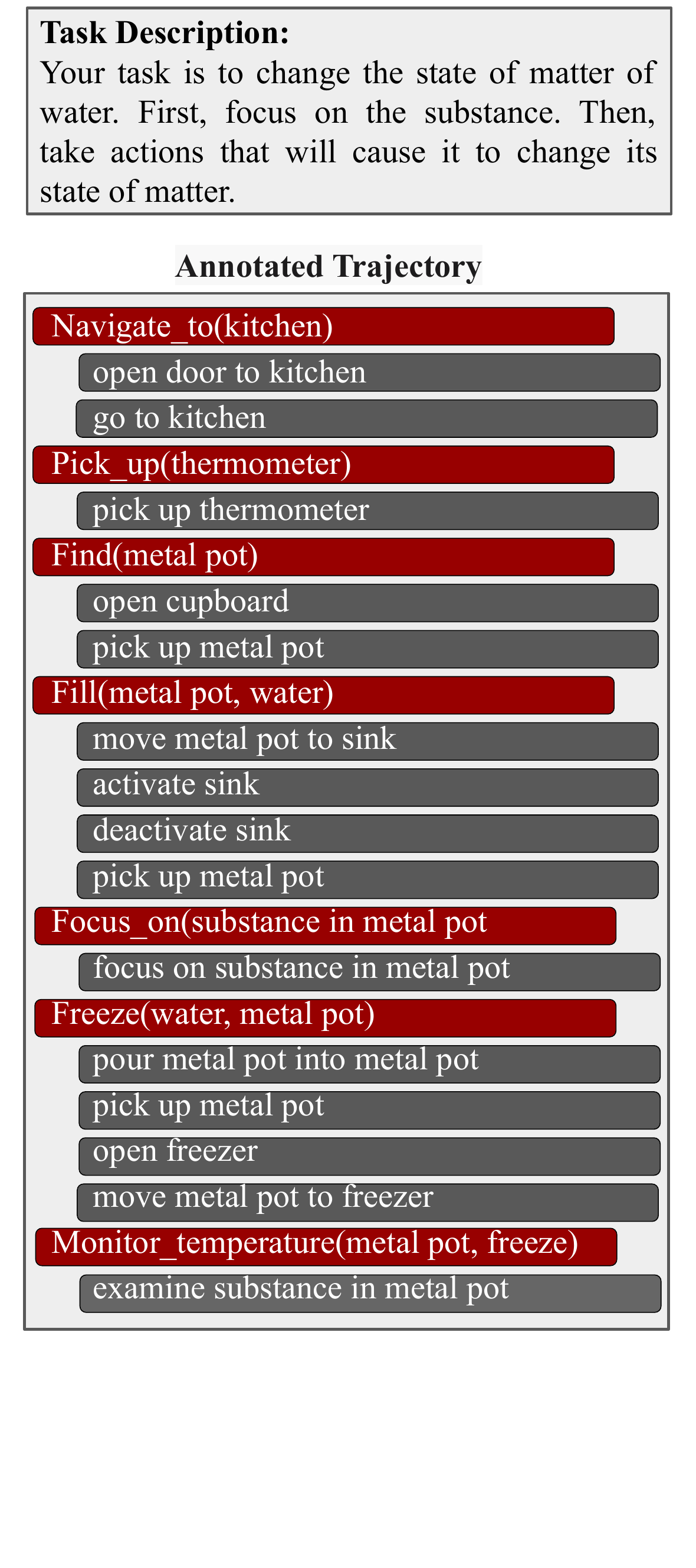}
  \caption{Example of annotating an expert trajectory with sub-goals for a particular variation of task 1-4 (\textit{change-the-state-of-matter-of}).
  Looking only at the original trajectory (i.e., ignoring the red rows), we gather the expert ended up changing the state of \textit{water} to be frozen. The expert had to navigate to the kitchen, find a thermometer and a metal pot, pour water into the pot, place it in the freezer, and continually monitor its temperature until frozen. Each of those milestones (highlighted in red) can be considered a sub-goal, encompassing a sequence of actions. Sub-goals can be shared across different tasks, facilitating generalization. We opted for a format that looks like function calls to encourage reusability (e.g., \textit{fill(metal pot, water)}).
  }
  \label{fig:example_1}
\end{wrapfigure}

Distilling knowledge from LLMs offers significant advantages, allowing for the training of specialized local models rather than depending on an LLM as a general model. This approach not only enhances privacy, particularly for systems with security-sensitive considerations like co-pilot models, but also provides greater flexibility in tailoring models for specific tasks. Additionally, the use of a smaller model offers the advantage of versatility across diverse applications without size constraints, including device models and mobile apps.
Another challenge with LLMs is their susceptibility to hallucinations. This tendency poses a hindrance to their effective execution of long-tail planning, especially in interactive decision-making scenarios. 

In our research, we leverage the knowledge of LLMs to train an autonomous agent for effective decision-making in complex interactive text environments, utilizing small language models as our policy. Knowledge Distillation facilitates the training of smaller policies, allowing seamless integration of LLM knowledge. To address the challenges at hand, adopting a two-level planning approach proves beneficial for reducing hallucination -- one for high-level reasoning to formulate sub-goals and another for low-level action planning to execute each sub-goal. 

\Cref{fig:example_1} illustrates this concept in the task of freezing water from ScienceWorld~\citep{wang2022scienceworld}. The agent's subtasks involve navigating to the kitchen, finding a thermometer and a metal pot, pouring water into the pot, placing it in the freezer, and continuously monitoring its temperature until frozen. These constitute sub-goals generated by a high-level model, with each sub-goal subsequently executed by a low-level model. The generation of sub-goals empowers an autonomous agent to expedite learning for the current task and reuse similar sub-goals in various tasks to have more generalization.

\vspace{0.2cm}
The contributions in this work are:
\begin{itemize}[leftmargin=*]
    \item We employ Knowledge Distillation from an LLM to train a high-level policy capable of generating sub-goals without making assumptions about the specific set of sub-goals. Notably, these sub-goals remain flexible, accommodating various sequences of actions.
    \item We demonstrate that employing Knowledge Distillation with hierarchical policies surpasses the performance achieved by both standalone imitation learning and its combination with in-context learning.
    \item We illustrate that this approach is more cost-effective in terms of the number of calls to an LLM compared to other methods utilizing in-context learning.
    \item We introduce an effective approach instead of using computational requirements of LLM and their restricted number of calls for using in interactive decision making tasks.
\end{itemize}


\section{Related Work}



\paragraph{Using LLMs for Action Planning}
Recent works have demonstrated the ability of LLMs to perform action planning for interactive decision making process without any additional training~\citep{huang2022language}.
ReAct~\citep{yao2022react} proposes a way of prompting an LLM with interleave reasoning step and action taking step. That led the resolution of a variety of language-based reasoning and decision-making tasks. This approach empowers the model to construct high-level plans for effective action.
Reflexion~\citep{shinn2023reflexion} draws inspiration from  reinforcement learning, employing a framework to reinforce language agents through linguistic feedback. At the end of each trial, it uses self-reflection to determine what went wrong with the task and keeps it in a memory. Then it leverages this information for the next trial. 

Some works use a programmatic LLM prompt structure with available actions and objects in an environment to translate natural language commands into robot policy code via few-shot examples~\citep{liang2023code, singh2023progprompt}. 
\citet{khot2022decomposed} introduced a decomposed prompting approach wherein a task is broken down into simpler sub-tasks, allowing for recursive handling. Subsequently, these sub-tasks are assigned to sub-task-specific LLMs, with both the decomposer and the sub-task LLMs with their own few-shot prompts. \citet{sun2023pearl} uses three steps, action mining, plan formulation, and plan execution to decompose a question into a sequence of actions by  few-shot prompting of LLMs. In~\citet{prasad2023adapt} tasks are decomposed explicitly by a separate LLM through prompting when an executor is unable to execute a given sub-task.


\paragraph{Imitation learning}
Some works employ imitation learning to train a language model as the agent's policy, as seen in offline decision transformers~\citep{torabi2018behavioral}. The inputs consist of states, actions, and reward-to-go values, which are fed into a transformer. This transformer then predicts actions in an autoregressive manner, utilizing a causal self-attention mask~\citep{chen2021decision}.
Contextual Action Language Model (CALM)~\citep{yao2020keep} is another work which uses a fine-tuned language model with oracle data to generate a set of candidate actions which are then passed to a policy network to select the best one.
In~\citet{nakano2021webgpt}, the authors fine-tune GPT-3 to address long-form questions within a web-browsing context. Human feedback is employed as a direct optimization measure for enhancing the quality of answers generated by the model.

\paragraph{Knowledge Distillation:}
Knowledge Distillation (KD) typically falls into two categories: black-box KD and white-box KD. In black-box KD, only the teacher's predictions are available for guidance, while in white-box KD, we have access to the teacher's parameters~\citep{gou2021knowledge}.
Recently, black-box KD has gained widespread use for fine-tuning original models using self-instruct techniques, as proposed by \citet{wang2022self}, or for smaller models~\citep{taori2023alpaca, chiang2023vicuna, peng2023instruction} through the utilization of prompt-response pairs generated by LLMs.
\citet{west2021symbolic} introduces symbolic KD from text rather than logits. This process involves the transfer of knowledge from a large, general model to a more compact commonsense model, facilitated by a commonsense corpus, yielding a commonsense knowledge graph and model.
The work by~\citet{hsieh2023distilling} trains a smaller model that outperform LLM using reasoning steps called rationales. They incorporated rationales as informative supervision to train smaller models with less training data.

 
\paragraph{Complex interactive text environments}
In text-based games, an agent interacts with the environment by reading and writing text while aiming towards the end game or solving a given task. Out of the recent frameworks that deals with generating and interfacing text-based games~\citep{cote18textworld, hausknecht19, ALFWorld20, murugesan2021textbased}, we use ScienceWorld~\citep{wang2022scienceworld} which is very complicated by having a large set of objects, actions, and tasks.

\begin{figure}[t!]
  \includegraphics[clip, trim=0.5cm 14cm 0.5cm 1.5cm, width=\textwidth]{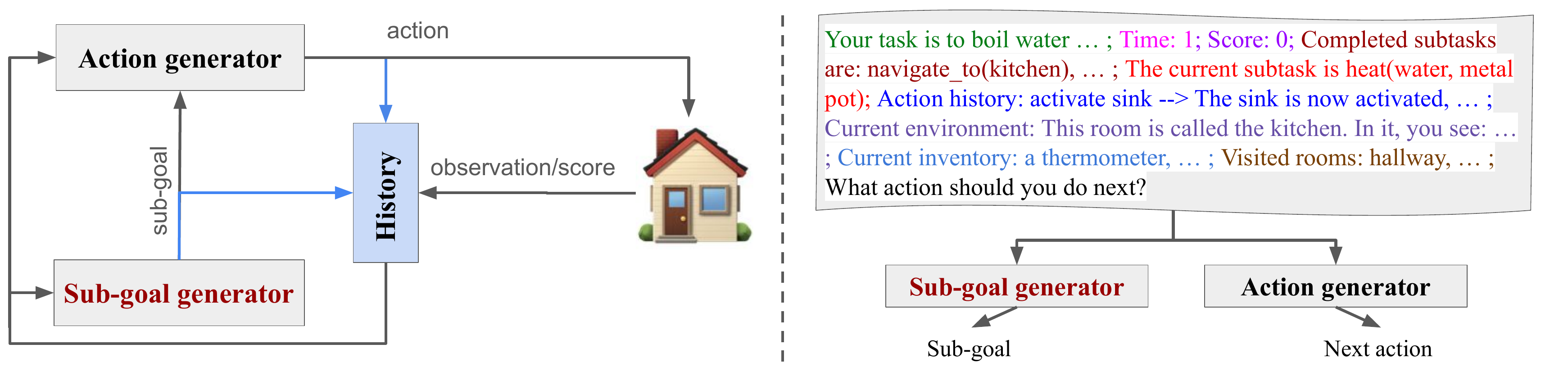}
  \caption{
  On the left, a schematic view of our approach is shown. There are two modules: the sub-goal generator and action generator. The sub-goal generator provides a sub-goal for the action generator, which predicts the next action given the current sub-goal and history. On the right, the inputs and outputs of both modules are illustrated. The input comprises different parts including task description, completed sub-goal, current sub-goal, a history of recent actions-observations, and more, each highlighted in a distinct color.}
  \label{fig:schematic}
\end{figure}

\section{Model} 

In this paper, we propose to train a hierarchical policy by combining KD from an LLM and imitation learning from expert trajectories. This section describes both modules in detail and we refer the reader to~\Cref{fig:schematic} for a schematic view. We first formulate the problem as a POMDP (\Cref{sec:pomdp}). 
Next, we describe what knowledge we are distilling from an LLM to guide the agent in accomplishing tasks (\Cref{sec:kd}). Then, we detail how both the high-level and low-level policies of the hierarchical policy are trained (\Cref{sec:hierarchical-policy}).


\subsection{Problem Formulation}
\label{sec:pomdp}

ScienceWorld~\citep{wang2022scienceworld} can be defined as a partially observable Markov decision process (POMDP), where observations provide information solely on environmental changes induced by the current action. ScienceWorld is an interactive text environment meaning all task instructions, observations and actions are expressed in textual form. 
Importantly, both observations and rewards in this environment are conditioned by the ongoing task.

Given a language vocabulary $V$ and an arbitrary maximum number of tokens $N$, an observation is defined such as $o \in \Omega  \subset V^N$, a reward such as $r \in \mathrm{R}$ and an action as $a \in A \subset V^N$. Finally, a task or goal description is shown by $g \in G \subset V^N$.

We formalize the problem as a goal-augmented POMDP $M = \left ( S, V, A, \Omega, G, T, R, O, \gamma \right )$ with $S$ the state space, $A \subset V^N$ the action space, $\Omega \subset V^N$ the observation space, $G \subset V^N$ the goal space, $T : S \times A \times G \rightarrow S$ the goal-conditioned transition function, $R : S \times A \times G \rightarrow \mathrm{R}$ the goal-conditioned reward function, $O : S \rightarrow V^N$ an (unknown) observation function mapping a state to a textual description and $\gamma$ the discount factor. We assume $\gamma = 1$ in our experiments. 

\subsection{Distilling Knowledge from an LLM}
\label{sec:kd}

The initial step in training our policies is creating a dataset. This dataset should include sub-goals along with their corresponding aligned sequences of actions for each task.

To generate sub-goals along with their corresponding aligned sequences of actions we do the following steps. We assume access to a collection of expert trajectories. Then we prompt an LLM with two in-context examples. Each example is composed of a task description, a similar task as the one we wish to annotate, and its expert trajectory. The example also contains a set of sub-goals, with the sequences of actions linked to each sub-goal.

Given the two in-context examples and a new task description with its expert trajectory, the LLM is then instructed to generate a response. The response is a set of sub-goals with their associated list of actions. The generated list of actions is used to determine each sub-goal corresponds to which segment of the expert trajectory.
It is important to note that these responses are collected only for the training tasks for which we assume having access to expert trajectories.
Also, it is important to point out that the LLM is not generating any novel trajectories.

\begin{figure*}[t]
  \includegraphics[clip, trim=0.0cm 18cm 6cm 0cm, width=\textwidth]{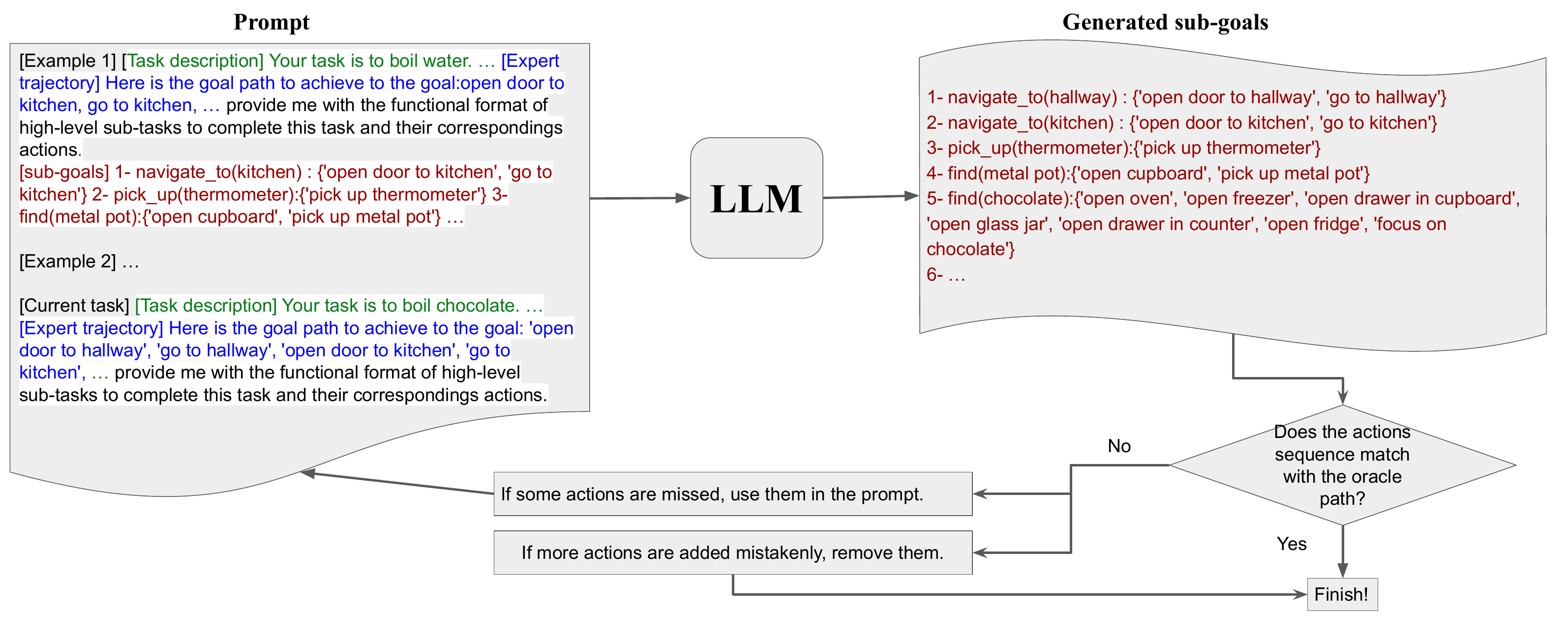}
  \caption{The figure demonstrates KD to generate sub-goals using an LLM. The LLM is presented with a prompt containing two in-context examples. Each example is composed of a task description in green and an expert trajectory detailing the steps to accomplish that task in blue. It also includes the expected set of sub-goals with their corresponding sequences of actions in red. Following this, we provide a new task description and trajectory, and we let the LLM generate the associated sub-goals and segmented actions.
  }
  \label{fig:KD}
\end{figure*}

\Cref{fig:KD} illustrates the prompt examples for task $1-1$ which is boiling a given substance. To ensure more uniform sub-goals that can generalize across tasks, we opted for a format that looks like function calls. Since that format was shown in the in-context examples, the LLM-generated sub-goals mimic this format as well making them easier to parse.

\begin{figure*}
  \includegraphics[width=0.85\textwidth]{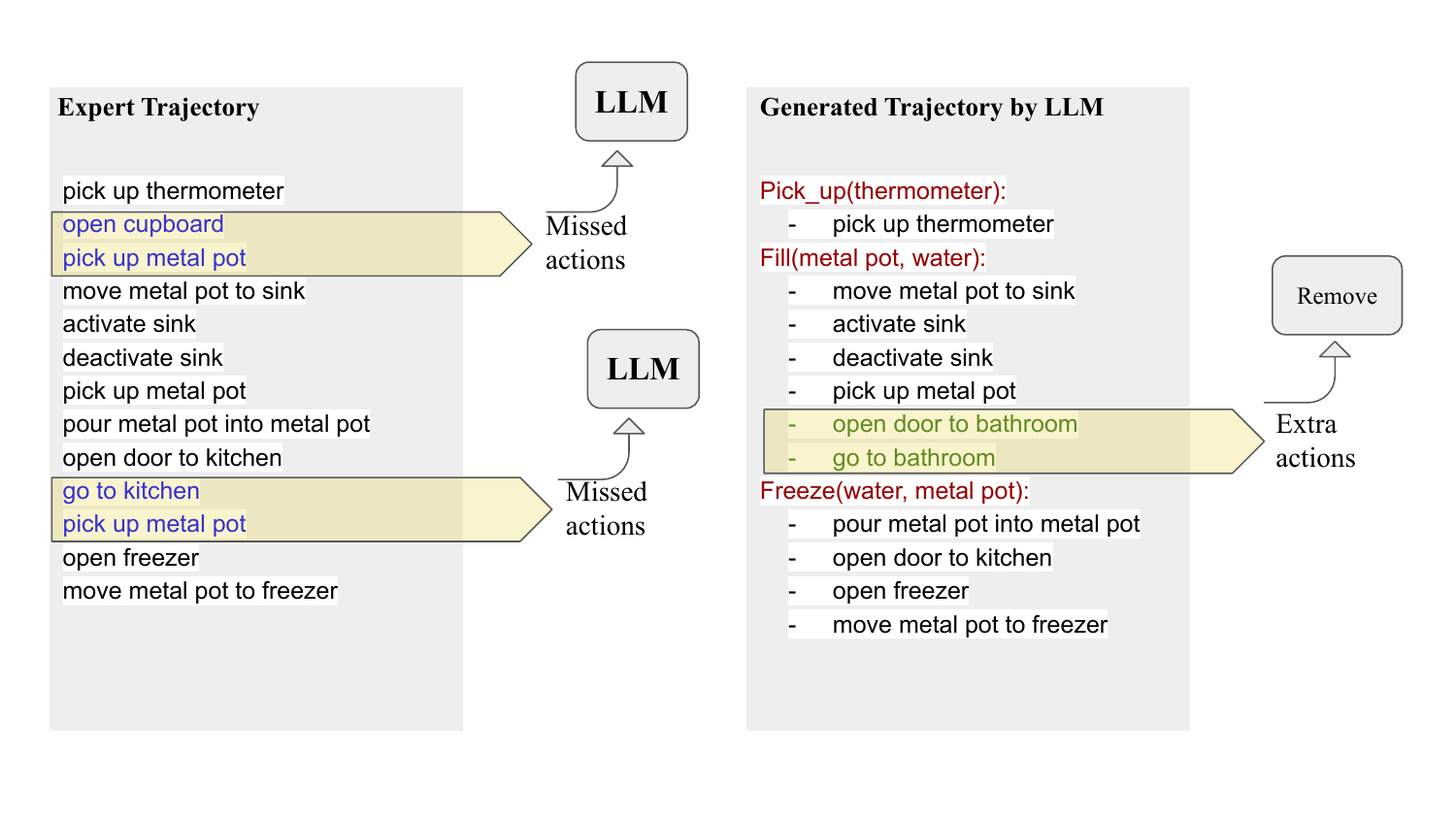}
  \caption{Example of a trajectory generated by the LLM deviating from the provided expert trajectory. In this example, which is for a boiling task, certain actions are omitted in the generated trajectory, indicated in blue in the left box. To address these missing actions, we group them into sequences and prompt the LLM to generate sub-goals for them. If the generated trajectory includes additional actions, such as the green actions in the right box, we simply remove them to align with the expert trajectory.
  }
  \label{fig:match_action}
\end{figure*}

Since the expert trajectories for some tasks can be long ($+100$ actions), the generated sub-sequence of actions corresponding to each sub-goal may not align exactly with the expert trajectory. Sometimes, it might miss certain actions, while in other instances, it might include additional actions, especially when there are repeated actions in the trajectory. To address this, we use a trajectory alignment process that finds the minimal set of edits to go from the generated trajectory to the expert trajectory according to the Levenshtein distance. For each “remove” edit, i.e. the generated trajectory has superfluous actions, we simply remove those from the generated trajectory. On the other hand, for “add'' edit, i.e. the generated trajectory is missing some actions, we prompt the LLM to generate a new sub-goal for those. An example is shown in~\Cref{fig:match_action}.



In the resulting annotated dataset, each data point follows the same format as used by ~\citet{lin2023swiftsage} but with the added mention of completed sub-goals and the current sub-goal. Precisely, it corresponds to:
\begin{itemize}
    \item \textbf{Input}: task description, number of steps, current score, completed sub-goal, current sub-goal, a history of $10$ recent actions-observations, current items in the room, inventory, and the visited rooms.
    \item \textbf{Target}: next action, next sub-goal.
\end{itemize}


\subsection{Hierarchical Imitation Learning}
\label{sec:hierarchical-policy}
With the dataset obtained from distilling knowledge from an LLM, we can now focus on training the policies.

\paragraph{Low-level policy:} 

The low-level policy is a language model (LM) which is trained through imitation learning using the annotated dataset. The goal is to have a model much smaller than an LLM so it can fit on a single machine and run faster, ideally below a billion of parameters.
This policy learns to predict the next action given the current task description, the 10 previous observation-action pairs, the previous completed sub-goals, and the current sub-goal.
We refer to this policy as the \textbf{{action generator}}.



\paragraph{High-level policy:} 

The high-level policy is another LM with a reasonable size. It is trained using the annotated dataset to generate the next sub-goal given the previous sub-goals and a short history, i.e. the last 10 actions and observations. So the high-level policy generates sub-goals while the low-level policy generate actions. Moreover, this policy conditions on the same input information as for the action generator. We call this policy the \textbf{{sub-goal generator}}.

\paragraph{Hierarchical policy:} 
During inference, we first leverage the high-level policy to generate a sub-goal. This generated sub-goal is then fed into the action generator, allowing it to produce the next action aligned with the provided sub-goal. This sequential approach serves as a guiding cue for the action generator, particularly when the trajectory to achieve the goal is complex or long. Moreover, it serves to prevent the action generator from generating actions that might deviate the agent from the correct path, thereby improving the precision and relevance of the actions being generated.






\section{Experiments}

\subsection{Environment}
We chose ScienceWorld~\citep{wang2022scienceworld} as the environment due to its complexity and the diverse range of tasks it encompasses. 
This environment is an interactive multi-task text-based game where the agent conducts elementary science experiments in a simulated environment. Each experiment is designed as a separate task. For example, \textit{"Your task is to boil water. For compounds without a boiling point, combusting the substance is also acceptable. First, focus on the substance. Then, take actions that will cause it to change its state of matter"}. To complete a task, the agent must perform multiple actions and receives the result of each action as an observation and a score. The observations and actions are in text format. An observation describes the changes in the environment, and the score is a numerical value ranging from $0\%$ to $100\%$, indicating the degree of completion of the current task through the current action. 


Furthermore, ScienceWorld is a benchmark with 30 distinct tasks spanning $10$  science domains which are widely different (\Cref{app:task_det}). For instance, in the "Changes of State" task, the agent is required to locate and use heating/freezing sources to alter the state of a substance (e.g., ice or chocolate). Conversely, in a task such as "Mendelian Genetics," the agent is tasked with determining whether a specified trait (e.g., white flower color) is dominant or recessive in a plant. These examples illustrate the substantial diversity across the domains, ranging from physical transformations to genetic analyses, underscoring the broad spectrum of challenges within ScienceWorld.

On top of that, ScienceWorld has $10$ different locations, more than $200$ object types, and $25$ action templates which makes the search space very larger for the agent. 
Each type of task has different variations in which the task objects, the agent's initial location, and random contents of each room are altered.
 
\subsection{Experimental Setup}
The environment has separate sets of variations for train and test. In the test variations, the combinations of objects and conditions are not seen in the train set. Following the experimental setup in \citep{lin2023swiftsage}, if the number of variations is more than $10$, we consider only the first $10$ variations. 

Our base models for both policies is a pre-trained \textsc{flan-t5-large}~\citep{chung2022scaling} with 700M parameters. For the both polices, we used greedy decoding at inference. We also conduct an ablation study over different model sizes (\Cref{fig:scale}). For fine-tuning the policies, we use all the training tasks and their variations ($3600$ games in total) from ScienceWorld. We vary the number of training epochs in function of the size of the models (see \Cref{sec:app-finetunning}).

\subsection{Baseline Agents}
We compare our approach with other works that leverage LLMs. Some rely only on prompting such as SayCan, ReAct, and Reflexion, but SwiftSage also do imitation learning. Here is a brief description of each method.


\paragraph{SayCan:} the LLM initially offers a set of actions along with their respective ranks. Then, a value-based method is employed to re-rank these actions in order to determine the most rewarding action for execution~\citep{brohan2023can}. 
\paragraph{ReAct:} the LLM generates actions by incorporating the provided prompt and the history of generated texts. It employs reasoning traces as intermediate thought steps during the action generation to refine a plan for the upcoming steps~\citep{yao2022react}.
\paragraph{Reflexion:} the language agent reflects the task feedback at each trial in the form of text and retains this information within an episodic memory. During the subsequent trial, it leverages the stored memory text to enhance its decision-making process~\citep{shinn2023reflexion}.
\paragraph{SwiftSage:} this method comprises two components: Swift, a fine-tuned LM to predict actions, and Sage, a module that queries an LLM for planning when the performance of Swift is inadequate (as determined by some handcrafted rules)~\citep{lin2023swiftsage}. 
\paragraph{Swift-only:} this is the Swift part of the SwiftSage method which only has the fine-tuned LM to predict the actions. We consider this method as a strong baseline and the most comparable to our approach as it relies on imitation learning without the need for querying an LLM during inference. 


Note that all baselines use ChatGPT (GPT-3.5) as their LLM.

\begin{table}[t]
  \centering
  \caption{The table illustrates the overall average score (\%) across all test tasks on the ScienceWorld benchmark for SayCan, ReAct, Reflexion, Swift-only, SwiftSage, and our algorithm (last column). The \textit{Solved Task Types} row represents the number of task types for which an agent manages to solve all the test variations. The table also shows the average scores for tasks with a short, medium, and long length of expert trajectory. The rows \textit{Task 1-1} and \textit{Task 3-3} display the scores for each of them in which our approach does not work well in comparison with the other methods. The $^*$ denotes scores reported from~\citep{lin2023swiftsage} which all use ChatGPT (GPT-3.5).
  }
\begin{tabular}{|c||c c c c c c|}
  \hline
   Methods &  SayCan$^*$ & ReAct$^*$ & Reflexion$^*$ & Swift-only & SwiftSage$^*$ & Ours  \\
   \hline
   \hline
    Overall Average &  25.22 & 19.76 & 23.40 & 46.25 & 62.22 &  65.43  \\
    Solved Task Types  &  0/30 & 0/30 & 4/30 & 4/30 & 2/30 & 11/30  \\
    \hline
    Short$^\dag$  &  37.24 & 28.95 & 39.19 & 79.68 & 72.81 &  91.61 \\
    Medium  & 20.06 & 21.09 & 14.73 & 35.80 & 55.34 &  62.83  \\
    Long  &  18.66 & 11.23 & 16.27 & 25.36 & 57.99 & 45.35   \\
    \hline
    Task 1-1 & 33.06 & 3.52 & 4.22 & 15.0 &  58.0 & 16.22  \\
    Task 3-3 & 99.56 & 76.19 & 72.54 & 59.5 & 66.9 & 5.6  \\
    \hline
  \end{tabular}
  \label{tab:1}
\end{table}

\subsection{Results and Analysis}

\paragraph{Main Results:}~\Cref{tab:1} compares the performance of the baselines with our approach in the ScienceWorld. 
The score for each task type is the average score (in percent) obtained for $10$ test variations. 
Our approach demonstrates an overall performance of $\textbf{65.43\%}$, surpassing Swift-only by 16.71\% (33.9\% relative increase), and showing a slight improvement over SwiftSage of 3.3\% (5.3\% relative). Interestingly, our method is able to solve all test variations (i.e., gets an average score of 100\%) for 11 out of the 30 task types. In contrast, SwiftSage solves them only for 2 task types, and Swift-only, only for 4 task types.

Additionally, we measured the performance of the agents with respect to the length of the tasks (a proxy for task complexity). The length of a task is determined by how many actions was needed by the expert to solve it.\footnote{Expert trajectories for test tasks were not seen during training.}
Following~\citet{lin2023swiftsage}, we group the tasks into three categories: \textit{Short} when the length is less than $20$ actions, \textit{Medium} when it falls between $20$ and $50$ (inclusively), and \textit{Long} if above $50$. As shown in \Cref{tab:1}, our approach outperforms other methods on short and medium tasks. 
On long tasks, we outperform all methods except SwiftSage, which has a substantial advantage here: The longer the task, the higher the chance it triggers one of the rules for Sage to take over.

As part of the comparison, there are other approaches that do not use a LLM including DRRN~\citep{he2016deep}, KG-A2C~\citep{ammanabrolu2019graph}, CALM~\citep{yao2020keep}, BC~\citep{torabi2018behavioral}, TDT~\citep{chen2021decision}. The results from \citep{wang2022scienceworld} show these approaches perform poorly, below $17 \%$, in ScienceWorld. For this reason, we did not include them here and only focus on approaches comparable with us. 

A key motivation for our approach is cost-effectiveness in terms of LLM queries. During training, we make one query to ChatGPT per task to identify the sub-goals within an expert trajectory. Sometimes mismatches occur between the expert trajectory and the actions assigned to each sub-goal by ChatGPT. When that is the case, we employ dynamic programming, with a maximum of $10$ attempts per task. This contrasts with other baseline methods, where LLM is queried for each action, incurring considerably higher costs.

\paragraph{Why is it failing on some task types?}

The performance of our algorithm in some tasks are low, (see~\Cref{app:table_diff_methods}).
In~\Cref{tab:1}, the scores of two tasks are presented.
One contributing factor is the variations in the test are very different from those in the training. For instance, the objects might be very different or the path to complete the task is very different and longer. The main culprit is the sub-goal generator which is not able to generate good sub-goals.

As a concrete example (\Cref{tab:bad_examples}), in the test variations for task 3-3, the agent needs to go to kitchen and then fill a jug with water. When looking at the transcript, we see the agent is able to go to kitchen but then when it arrives, the sub-goal generator issues a sub-goal which is not relevant, {\tt{FocusOn(fountain)}}. The agent attempts to focus on the fountain which is a wrong action and the game terminates with a score of 0. 

Another example is task 1-1 (\Cref{tab:bad_examples}) in which the agent should boil a substance. It should first find the substance but since the substance is in a totally different location than those seen during training, the sub-goal generator is not able to generate a good sub-goal for this step. Consequently the agent will do other actions and exhaust all the allocated time steps.

\definecolor{burgundy}{rgb}{0.5, 0.0, 0.13}
\begin{table}[ht]
\small
\centering
\begin{tabular}{|l|l||l|l|}
\hline
 \multicolumn{2}{|c||}{ \cellcolor[gray]{.7} \textbf{Example (task 3-3)}} &  \multicolumn{2}{|c|}{\cellcolor[gray]{.7} \textbf{Example (task 1-1)}}\\
 \hline
 \cellcolor[gray]{.9}\textbf{With Sub-goal}& \cellcolor[gray]{.9}\textbf{Expert Trajectory} & \cellcolor[gray]{.9}\textbf{With Sub-goal}&\cellcolor[gray]{.9} \textbf{Expert Trajectory} \\
 \hline
 \textcolor{burgundy} {NavigateTo(kitchen) }& -  & \textcolor{burgundy}{NavigateTo(kitchen)}&- \\
 - go to art studio & - go to art studio & - go to art studio&- go to art studio \\
 - go to outside & - go to outside & - go to outside&- go to outside \\
 - go to kitchen & - go to kitchen & - go to hallway&- go to hallway \\
 \textcolor{burgundy} {FocusOn(fountain)} & -  & \textcolor{burgundy}{NavigateTo(bedroom)}&-  \\
 \textcolor{red} {-focus on fountain} & \textcolor{black} {- move jug to sink} & \textcolor{red} {-go to bedroom}&\textcolor{black} {- go to workshop} \\
 & \textcolor{black} {- activate sink} & &\textcolor{black} {- pick up metal pot \newline containing gallium} \\
 &\textcolor{black} {- deactivate sink} & &\\
 &\textcolor{black} {- pick up jug} & &\\
 \hline
\end{tabular}
\caption{
Two instances where the performance of our algorithm is low. The first column displays the trajectory generated with sub-goals, while the second column presents the expert trajectory. Sub-goals are highlighted in dark red, accompanied by their corresponding actions, and incorrect actions are marked in red.}
\label{tab:bad_examples}
\end{table}

































\begin{figure}
     \centering
     \begin{subfigure}[b]{0.45\textwidth}
         \centering
         \includegraphics[width=\textwidth]{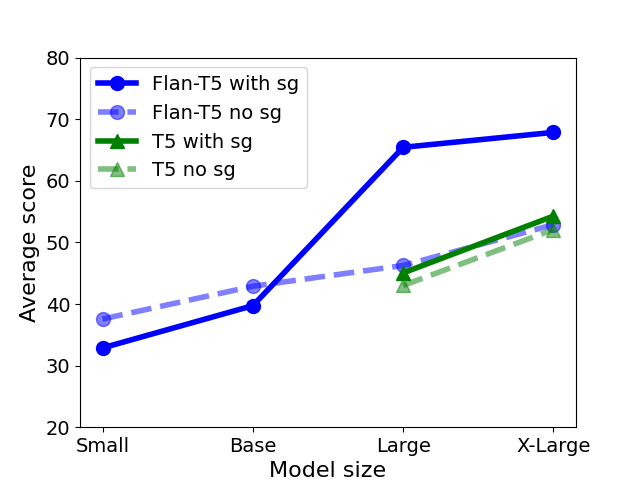}
          \caption{}
          \label{fig:scale}
     \end{subfigure}
     \hfill
     \begin{subfigure}[b]{0.45\textwidth}
         \centering
         \includegraphics[width=\textwidth]{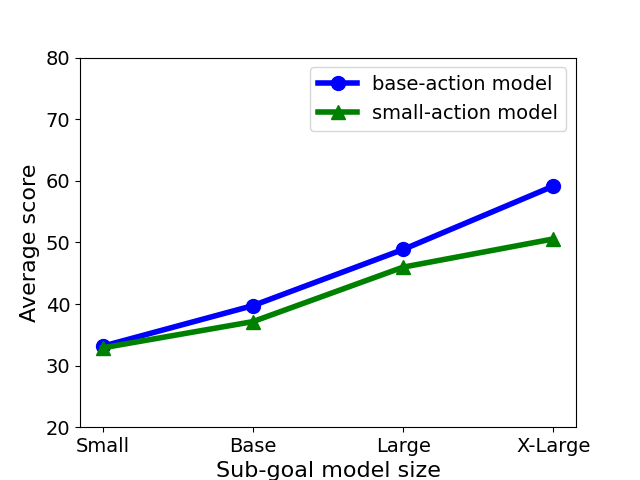}
          \caption{}
          \label{fig:sg_up}
     \end{subfigure}
\caption{a) Average scores across different model sizes for \textsc{flan-t5} and \textsc{t5}. For \textsc{t5} model, X-Large refers to \textsc{t5-3b}. The larger models work better and \textsc{flan-t5} performs also better than \textsc{t5}. Dashed lines represent models that are not conditioning on any sub-goals (``\textit{no sg}'') and equivalent to Swift-only. b) Average scores across different sizes of sub-goal generator while the action generator is kept to be base (blue) or small (green). Having larger sub-goal generators can significantly boost performance of small action generators.}
\label{fig:three graphs}
\end{figure}

\paragraph{The impact of scale:}
We conduct a comparison across various sizes of language models such as \textsc{flan-t5-xl}, \textsc{flan-t5-base}, and \textsc{flan-t5-small}. Additionally, we evaluate \textsc{t5-3b} and \textsc{t5-large} to determine the effectiveness of \textsc{flan-t5} versus \textsc{t5}. The results are illustrated in~\Cref{fig:scale}. In our initial findings, we observed that \textsc{flan-t5} outperforms \textsc{t5} significantly. Moreover, our results reveal a positive correlation between the LM size and its performance -- larger models generally yield better results. Intriguingly, we observe that for smaller models (\textsc{flan-t5-small} and \textsc{flan-t5-base}), not conditioning on sub-goals works slightly better than including them. This might be indicative that the sub-goal generator is not expressive enough to generate meaningful and effective sub-goals which in turn impacts the action generator policy and leads to lower scores.

\paragraph{The impact of sub-goals:}
To study the impact of the sub-goal generator's size on the overall performance, we try pairing different sizes of sub-goal generator while limiting the action generator to be small.
In~\Cref{fig:sg_up}, the average scores exhibit an upward trajectory. This can be attributed to the larger sub-goal generators producing more accurate and relevant sub-goals, subsequently empowering the action generator to generate more correct actions. See ~\Cref{app:table_diff_sizes} for a complete breakdown of the score per task type and per model size.

To further demonstrate the importance of the sub-goal, we generated random sub-goals and then fed them to the action generator. That yield an average score of $\textbf{6.4\%}$, indicating that the action generator do condition on the sub-goals, subsequently, it cannot solve the tasks effectively.
We conducted an additional experiment by altering the arguments of the sub-goals, as they have a functional format. If the argument corresponds to a location, we replaced it with another from the environment, and if it is an object, we replaced it with a randomly chosen object available at that step of the game. We named this approach \textit{semi-random} sub-goals. The result for this experiment is $\textbf{14.2\%}$, showing an increase in performance compared to the random sub-goals. 
Table \ref{tab:rand_sg} shows the average scores and \Cref{table:rand_sg_appendix} shows the score for each task.

\paragraph{Recovery from noisy sub-goals:}
We also assess the performance when both the action and sub-goal generators have been exposed to noisy sub-goals. More specifically, we consider two settings: applying noise 1) only at the first step, or 2) every 10 steps.
In the first setting, the first sub-goal is (semi-)randomly selected, while the subsequent sub-goals are generated using the \textsc{Flan-t5-large} sub-goal generator.
In the second experiment, a sub-goal is (semi-)randomly selected every $10$ steps instead of using the sub-goal generator for all steps. Table~\ref{tab:rand_sg} shows the overall scores for both settings and a breakdown per task types is presented in~\Cref{tab:ret_sg}.


In both scenarios, semi-random selection ($\textbf{53.1\%}$ and $\textbf{43.3\%}$) yields better results, as it closely resembles the sub-goals generated by the sub-goal generator. Some tasks achieve a score of $100$, indicating successful recovery from noisy sub-goals. While overall scores are lower compared to using the \textsc{Flan-t5-large} sub-goal generator, it is still higher than using Swift only in the first setting and closely approaching it in the second setting (\Cref{app:ret-sg}).

\paragraph{Generalization on heldout task types:}
We select one or two task types from each science domain (see highlighted ones in \Cref{app:table_tasks_descrip}) to train the action and sub-goal models. Then, we assessed their performance on the rest of the task types. We compared our algorithm against the Swift-only baseline. The average total scores are $\textbf{40.63\%}$ with sub-goals vs. $\textbf{36.56\%}$ for Swift-only. For unseen tasks, the scores are $\textbf{27.72\%}$ with sub-goals vs. $\textbf{15.25\%}$ for Swift-only. This suggests that using sub-goals helps improve generalization across unseen tasks. The scores for each task are presented in \Cref{tab:held_out}.




\begin{table}
    \centering
    \begin{tabular}{ccc|ccc}
      \toprule 
      \multicolumn{3}{c|}{\textbf{Random}} & \multicolumn{3}{c}{\textbf{Semi-random}} \\
      first & 10 steps & each & first & 10 steps & each \\
      \midrule
      39.1\% & 37.6\% & 6.4\% & 53.1\% & 43.3\% & 14.2\% \\
      \bottomrule
    \end{tabular}
    \caption{
    Average performance for randomly generated sub-goals. Sub-goals are selected randomly (or semi-randomly) at either the \textbf{first} step, every \textbf{10 steps}, or \textbf{each} step.}
    \label{tab:rand_sg}
\end{table}

\section{Discussion and Limitation}


In contrast to SwiftSage, which relies on interactive usage of the ChatGPT API to handle planning, our approach makes use of a trained sub-goal generator  to guide the action generator. Moreover, our framework empowers the agent to retrieve a nearly optimal trajectory by supplying the appropriate sub-goal. Nevertheless our framework has significantly reduced the frequency of API calls, which are both expensive and not universally accessible.
ReAct, Reflexion, and SwiftSage require human annotations to correct sub-goals and predict a reasonable action. However in our approach, we do not need human help to predict sub-goals or provide precise prompts.


\paragraph{Generalization:}
In this work, our focus is on optimizing performance within the environment, and there might be a potential limitations when transitioning to entirely different scenarios.
If we test it in a distinct environment, the performance may not be optimal, given the fine-tuning with data specific to the ScienceWorld environment. It's acknowledged that for generalization across diverse scenarios, an LLM may perform better, given its capacity to handle a broader range of inputs and contexts. 

\paragraph{Goal Modification:}
When the agent encounters challenges in solving the current sub-goal, it will often find itself cycling through the same sub-goal for several steps. Consequently, the action generator repeats a sequence of actions mirroring recent ones. Sometimes the sub-goal generator will adjust the sub-goal slightly based on the input and that can be enough to get unstuck.

Ideally, we would like to avoid being stuck for several steps and learn to modify the sub-goal in the right way. One strategy involves online learning, where the controller is updated based on the reward from the environment. However, this approach carries the risk of catastrophic forgetting, necessitating additional measures such as loss modification and regularization to mitigate this risk. Another approach could involve incorporating an LLM alongside the controller. If the controller fails to produce effective actions, the LLM can suggest alternative sub-goals. This might have the risk of poor sub-goals and hallucinations which rewards might help but it is still challenging in such a sparse environment.

\section{Conclusion}
We introduce a straightforward yet highly effective approach for tackling complex text-based environments. Our framework leverages the knowledge of an LLM to extract sub-goals. A hierarchical policy of two LMs proposed: a high-level policy predicts a sub-goal, and a low-level policy, by using the predicted sub-goal, generates elementary actions. Through extensive experiments across $30$ task types in ScienceWorld, our approach demonstrates increase performance compared to state-of-the-art baselines, including standard imitation learning and SwiftSage.

As future directions for this work, we aim to delve into further exploration of goal modification strategies when the agent encounters challenges in solving the current sub-goal. This could involve breaking down or transforming a sub-goal into a more achievable form. Another venue for future research involves extending this approach to a multi-module environment. In such scenarios, the sub-goal generator could leverage each module as an independent source to generate diverse and context-specific sub-goals. Exploring strategies for goal modification and online learning is another avenue we are keen to pursue.

\subsubsection*{Acknowledgments}
Special thanks are due to Prasanna Parthasarathi for his invaluable insights, thoughtful brainstorming, and engaging discussions in the project. Also, thank you to Xingdi Yuan for initial discussions around knowledge distillation with LLM. Sarath Chandar is supported by the Canada CIFAR AI Chairs program, the Canada Research Chair in Lifelong Machine Learning, and the NSERC Discovery Grant. This research was enabled mostly by compute resources provided by Mila (mila.quebec) and partially by Microsoft.

\bibliography{collas2024_conference}
\bibliographystyle{collas2024_conference}

\newpage
\appendix
\section{Appendix}

\subsection{Few-shot Prompt for the Large Language Model }
\label{sec:app-prompt-gpt}
We employed the ChatGPT API as the large language model in our study.  The structure of the ChatGPT prompt is comprised of three main components. Firstly, there is a general description of the environment. The second part includes two examples, each containing the task description, an expert trajectory, and a set of sub-goals with their corresponding action sequences. Lastly, the prompt presents a new task description, along with an expert trajectory, then we ask the LLM to generate sub-goals for this new task. 

Here is the first part:
\begin{tcolorbox}[
enhanced,
left=2pt, right=2pt, top=2pt, bottom=2pt,
title=Description of the Environment]
You are a helpful assistant. You are in a simulated environment as an agent. A task and its description will be given to you. Suggest the best actions the agent can take based on the things you see and the items in your inventory to complete the task. Only use valid actions and objects. If you know what are around, then suggest the following actions. You are allowed to do the following actions with the objects. Open or close OBJ meaning open or close a container , Deactivate or activate OBJ meaning activate or deactivate a device, connect OBJ to OBJ meaning connect electrical components , disconnect OBJ meaning disconnect electrical components , use OBJ [on OBJ] meaning use a device/item , look around meaning describe the current room, look at OBJ meaning describe an object in detail, look in OBJ meaning describe a container’s contents, read OBJ meaning read a note or book, move OBJ to OBJ meaning move an object to a container, pick up OBJ meaning move an object to the inventory, put down OBJ meaning drop an inventory item, pour OBJ into OBJ meaning pour a liquid into a container , dunk OBJ into OBJ meaning dunk a container into a liquid , mix OBJ meaning chemically mix a container , go to LOC meaning move to a new location , teleport to LOC meaning teleport to a specific room , eat OBJ meaning eat a food , flush OBJ meaning flush a toilet, focus on OBJ meaning signal intent on a task object, wait [DURATION] meaning take no action for some duration, task meaning describe current task, inventory meaning list agent’s inventory, OBJ means objects. LOC means location. There are 10 locations centered around a house theme. These are: kitchen, bathroom, workshop, art studio, greenhouse, outside, bedroom, living room, foundry.
\end{tcolorbox}

Here are the two examples for task $4-1$, which is \textit{find-living-thing}. This is the second part of the prompt:

\begin{tcolorbox}[
enhanced,
left=2pt, right=2pt, top=2pt, bottom=2pt,
title= Two-shot examples of Task 4-1 (find-living-thing)]
\textbf{Example 1}
{\color{darkgreen} [Task Description] Your task is to find a(n) living thing. First, focus on the thing. Then, move it to the red box in the kitchen.}

{\color{blue} [Expert trajectory] Here is the goal path to achieve to the goal: open door to greenhouse, go to greenhouse, open door to outside, go to outside, focus on dove, pick up dove, open door to kitchen, go to kitchen, move egg dove egg in inventory to red box
}

Based on the given goal path, provide me with the functional format of high-level sub-tasks to complete this task and their corresponding actions.

{\color{red} [sub-goals] 1- navigate\_to(greenhouse): \{'open door to greenhouse', 'go to greenhouse'\} 2- navigate\_to(outside): \{'open door to outside', 'go to outside'\} 3- Focus\_on(dove): \{'focus on dove'\} 4- pick\_up(dove): \{'pick up dove'\} 5- navigate\_to(kitchen): \{'open door to kitchen', 'go to kitchen'\} 6- move(dove egg, red box): \{'move dove egg in inventory to red box'\} }

\textbf{Example 2}
{\color{darkgreen} [Task Description] Your task is to find a(n) living thing. First, focus on the thing. Then, move it to the green box in the kitchen. }

{\color{blue} [Expert trajectory] Here is the goal path to achieve to the goal: open door to kitchen, go to kitchen, open door to outside, go to outside, focus on egg turtle, pick up egg turtle, open door to kitchen, go to kitchen, move egg turtle egg in inventory to green box
}

Based on the given goal path, provide me with the functional format of high-level sub-tasks to complete this task and their corresponding actions.

{\color{red} [sub-goals] 1- navigate\_to(kitchen): \{'open door to kitchen', 'go to kitchen'\} 2- navigate\_to(outside): \{'open door to outside', 'go to outside'\} 3- Focus\_on(egg turtle): \{'focus on egg turtle'\} 4- pick\_up(egg turtle): \{'pick up egg turtle'\} 5- navigate\_to(kitchen): \{'open door to kitchen', 'go to kitchen'\} 6- move(egg turtle, green box): \{'move egg turtle in inventory to green box'\} }

\end{tcolorbox}

The third part is just a new task description with an expert trajectory:

\begin{tcolorbox}[
enhanced,
left=2pt, right=2pt, top=2pt, bottom=2pt,
title= Request for sub-goal generating]
{\color{darkgreen} [Task Description] Your task is to find a(n) living thing. First, focus on the thing. Then, move it to the green box in the living room.}

{\color{blue} [Expert trajectory] Here is the goal path to achieve to the goal: open door to hallway, go to hallway, open door to greenhouse, go to greenhouse, open door to outside, go to outside, focus on baby baby beaver, pick up baby baby beaver, open door to greenhouse, go to greenhouse, open door to hallway, go to hallway, open door to living room, go to living room, move baby baby beaver in inventory to green box
}

Based on the given goal path, provide me with the functional format of high-level sub-tasks to complete this task and their corresponding actions.
\end{tcolorbox}

\subsection{Input Prompt for the Policies}
\label{sec:app-input-lm}

Here, we show the inputs and outputs for both the action generator and sub-goal generator.  We followed the format used by SwiftSage~\citep{lin2023swiftsage}, incorporating sub-goals and making minor textual adjustments accordingly.

\definecolor{bondiblue}{rgb}{0.0, 0.58, 0.71}
\definecolor{brass}{rgb}{0.71, 0.65, 0.26}
\begin{tcolorbox}[
enhanced,
left=2pt, right=2pt, top=2pt, bottom=2pt,
title= Input for the action generator]
{\color{darkgreen} [Task Description] Your task is to find a(n) living thing. First, focus on the thing. Then, move it to the red box in the kitchen;}

{\color{purple}Time: 4;}
{\color{purple}Score: 16; $</s>$}

{\color{red}Completed subtasks are: navigate\_to(greenhouse), navigate\_to(outside). The current subtask is Focus\_on(dove);$</s>$}

{\color{blue}Action history:  $<extra\_id\_4>$ look around (+0) --> N/A |  $<extra\_id\_3>$ go to greenhouse (+16) --> You move to the greenhouse. |  $<extra\_id\_2>$ open door to outside (+0) --> The door is already open. |  $<extra\_id\_1>$ go to outside (+0) --> You move to the outside. | $</s>$ }

{\color{violet}Current environment: This outside location is called the outside. Here you see:  | the agent | a substance called air | an axe | a blue jay egg | a butterfly egg | a dove egg | a fire pit  | a fountain (containing a substance called water) | the ground | a substance called wood | You also see: | A door to the foundry  | A door to the greenhouse  | A door to the kitchen  |  $</s>$;}

{\color{bondiblue}Current inventory: In your inventory, you see: | an orange |  $</s>$;}

{\color{brass}Visited rooms: hallway, greenhouse, outside $</s>$;}

What action should you do next? $</s>$

\end{tcolorbox}

\begin{tcolorbox}[
enhanced,
left=2pt, right=2pt, top=2pt, bottom=2pt,
title= Output for the action generator]
focus on dove
\end{tcolorbox}

\begin{tcolorbox}[
enhanced,
left=2pt, right=2pt, top=2pt, bottom=2pt,
title= Input for the sub-goal generator]
{\color{darkgreen} [Task Description] Your task is to find a(n) living thing. First, focus on the thing. Then, move it to the red box in the kitchen;}

{\color{purple}Time: 4;}
{\color{purple}Score: 16; $</s>$}

{\color{red}The previous subtasks are: navigate\_to(greenhouse). The current subtask is navigate\_to(outside) ;$</s>$}

{\color{blue}Action history:  $<extra\_id\_4>$ look around (+0) --> N/A | $<extra\_id\_3>$ go to greenhouse (+16) --> You move to the greenhouse. |  $<extra\_id\_2>$ open door to outside (+0) --> The door is already open. |  $<extra\_id\_1>$ go to outside (+0) --> You move to the outside. | $</s>$ }

{\color{violet}Current environment: This outside location is called the outside. Here you see:  | the agent | a substance called air | an axe | a blue jay egg | a butterfly egg | a dove egg | a fire pit  | a fountain (containing a substance called water) | the ground | a substance called wood | You also see: | A door to the foundry  | A door to the greenhouse  | A door to the kitchen  |  $</s>$;}

{\color{bondiblue}Current inventory: In your inventory, you see: | an orange |  $</s>$;}

{\color{brass}Visited rooms: hallway, greenhouse, outside $</s>$;}

What subtask should you do next? $</s>$ 

\end{tcolorbox}

\begin{tcolorbox}[
enhanced,
left=2pt, right=2pt, top=2pt, bottom=2pt,
title= Output for the sub-goal generator]
Focus\_on(dove)
\end{tcolorbox}




\subsection{Implementation Details}
\label{sec:app-finetunning}

We set the maximum number of steps per episode to $100$. Additionally, we implemented an alternative termination condition: if the scores remain unchanged over the last $50$ steps, we stop the episode. This prevents the agent from getting stuck in a repetitive loop of actions that do not yield any changes, such as navigating between rooms. We selected this threshold to ensure that it is not too restrictive, considering that certain tasks with lengthy trajectories may require a long sequence of actions to achieve a reward.

The learning rate in all of the experiments is $1e-4$. The values for $max\_source\_length=1024$ and for $max\_target\_length=30$. The batch size for training is $8$. 
We set the number of epochs to $20$ during training. In the evaluation phase, checkpoints were selected based on their loss values, encompassing the checkpoint with the lowest loss and the subsequent three checkpoints. The final choice for test was determined among these checkpoints, with priority given to the one demonstrating the highest score in the test set.

For both the action generator and sub-goal generator, we used greedy decoding.
When the generated actions is invalid we attempt to find the closes match from the list of admissible commands provided by ScienceWorld.


\subsection{Dataset Statistics}
\label{app:task_det}

In~\Cref{app:table_tasks_descrip}, we provide tasks' names and their variations for all of the tasks in ScienceWorld~\citep{wang2022scienceworld}. 
For each task, variations are partitioned into 50\% training, 25\% development, and 25\% test sets. In the development and test sets, variations include substances, animals, or plants that are not seen in the training.

\begin{table}
\centering
\begin{tabular}{clcc}
\hline
 Task Type  &Topic& Task Name& \# Variations\\
\hline
 \rowcolor{LigthGray}1-1  &Matter& boil& 30\\
 1-2  &Matter& melt& 30\\
 \rowcolor{LigthGray}1-3  &Matter& freeze& 30\\
 1-4  &Matter& change-the-state-of-matter-of& 30\\
 \rowcolor{LigthGray}2-1  &Measurement& use-thermometer& 540\\
 \rowcolor{LigthGray}2-2  &Measurement& measure-melting-point-known-substance& 436\\
 2-3  &Measurement& measure-melting-point-unknown-substance& 300\\
 \rowcolor{LigthGray}3-1  &Electricity& power-component& 20\\
 3-2  &Electricity& power-component-renewable-vs-nonrenewable-energy& 20\\
 \rowcolor{LigthGray}3-3  &Electricity& test-conductivity& 900\\
 3-4  &Electricity& test-conductivity-of-unknown-substances& 600\\
 \rowcolor{LigthGray}4-1  &Classification& find-living-thing& 300\\
 \rowcolor{LigthGray}4-2  &Classification& find-non-living-thing& 300\\
 4-3  &Classification& find-plant& 300\\
 4-4  &Classification& find-animal& 300\\
 \rowcolor{LigthGray}5-1  &Biology& grow-plant& 126\\
 5-2  &Biology& grow-fruit& 126\\
 \rowcolor{LigthGray}6-1  &Chemistry& chemistry-mix& 32\\
 6-2  &Chemistry& chemistry-mix-paint-secondary-color& 36\\
 \rowcolor{LigthGray}6-3  &Chemistry&  chemistry-mix-paint-tertiary-color&  36\\
 \rowcolor{LigthGray}7-1  &Biology&  lifespan-longest-lived&  125\\
 \rowcolor{LigthGray}7-2  &Biology&  lifespan-shortest-lived&  125\\
 7-3  &Biology&  lifespan-longest-lived-then-shortest-lived&  125\\
 \rowcolor{LigthGray}8-1  &Biology&  identify-life-stages-1&  14\\
 8-2  &Biology&  identify-life-stages-2&  10\\
 \rowcolor{LigthGray}9-1  &Forces&  inclined-plane-determine-angle&  168\\
 \rowcolor{LigthGray}9-2  &Forces&  inclined-plane-friction-named-surfaces&  1386\\
 9-3  &Forces&  inclined-plane-friction-unnamed-surfaces&  162\\
 10-1  &Biology&  mendelian-genetics-known-plant&  120\\
 \rowcolor{LigthGray}10-2 &Biology& mendelian-genetics-unknown-plant& 480\\
 \hline
\end{tabular}
\caption{ScienceWorld's tasks names and numbers of variations. The highlighted rows show training task types for the generalization experiment.}
\label{app:table_tasks_descrip}
\end{table}

\subsection{Scores of Each Task}

The average score of each task of the ScienceWorld is shown in~\Cref{app:table_diff_methods}. The methods are SayCan, ReAct, Reflexion, Swift-only, SwiftSage and our algorithm. In all of them ChatGPT is used as the LLM. The language models for Swift-only, SwiftSage and our algorithm are \textsc{flan-t5-Large}. For the methods, SayCan, ReAct, Reflexion, and SwiftSage we used the results from \citet{lin2023swiftsage}. However, we reproduced the results for Swift-only and we found a lower performance than what was reported in the paper. So here we presented our scores. 

The scores for SayCan, ReAct, Reflexion, and SwiftSage when they use GPT4 as the LLM are higher according to the results reported in~\citet{lin2023swiftsage}. However, due to limited access to GPT-4, we utilized ChatGPT, and thus, we present the results obtained using ChatGPT.


\begin{table}[ht]
  \centering
    \caption{The table illustrates the results for each task of the ScienceWorld for SayCan, ReAct, Reflexion, Swift-only, SwiftSage and our algorithm. Each row shows the average score of the test variations for a task type. Column \textit{Length} shows the average lengths of the expert trajectories. }
  \begin{tabular}{|c|c|cccccc|}
  \hline
   Task Type & Length & SayCan & ReAct & Reflexion & Swift-only & SwiftSage& Ours \\
   \hline
    1-1 & 107.7 &  0& 3.52 & 4.22 & 15.0 & 58.0 & 16.22 \\
    1-2 & 78.6 &  0& 13.70 & 10.61 & 24.4 & 58.5 & 0.2 \\
    1-3 & 88.9 &  0& 7.78 & 7.78 & 32.2 & 38.5 & 30.33 \\
    1-4 & 75.2 &  0& 9.88 & 0.92 & 57.4 & 62.5 & 65.0 \\
    2-1 & 21.4 &  1.7& 7.19 & 5.92 & 9.4 & 47.9 & 98.55  \\
    2-2 & 35.2 &  14.1& 6.10 & 28.59 & 6.7 & 53.3 & 46.0 \\
    2-3 & 65.0 &  93.7& 22.37 & 22.37 & 5.7 & 48.6 & 45.0 \\
    3-1 & 13.6 &  19.3& 56.0 & 100.0 & 70.0 & 72.7 & 100 \\
    3-2 & 20.8 &  8.7& 54.33 & 17.45 & 48.3 & 50.3 & 78.2 \\
    3-3 & 25.6 &  22.0& 76.19 & 72.54 & 59.5 & 66.9 & 5.6  \\
    3-4 & 29.0 &  36.4& 36.4& 70.22 & 69.0 & 78.1 & 46.0 \\
    4-1 & 14.6 &  11.7& 26.67 & 64.93 & 100.0 & 100.0 & 100 \\
    4-2 & 8.8 &  76.0& 80.0 & 87.27 & 100.0 & 97.5 & 100 \\
    4-3 & 12.6 &  11.4& 53.33 & 16.42 & 94.4 & 58.3 & 100  \\
    4-4 & 14.6 &  9.5& 27.50 & 100.0 & 100.0 & 100.0 & 100 \\
    5-1 & 69.5 &  11.3& 11.1& 5.8& 13.4 & 57.5 & 100 \\
    5-2 & 79.6 &  75.0& 18.8& 47.6& 44.6 & 50.9 & 65 \\
    6-1 & 33.6 &  13.5& 35.0& 22.4& 26.2 & 43.2 & 66.87 \\
    6-2 & 15.1 &  25.0& 20.0& 10.0& 53.3 & 63.3 & 100 \\
    6-3 & 23.0 &  58.4& 16.7& 40.0& 11.1 & 27.4 & 100 \\
    7-1 & 7.0 &  75.0& 37.5& 75.0& 83.3 & 75.0 & 100 \\
    7-2 & 7.0 &  100.0& 50.0& 75.0& 100.0 & 60.0 & 100 \\
    7-3 & 8.0 &  31.7& 31.7& 28.1& 77.8 & 68.3 & 100 \\
    8-1 & 40.0 &  5.6& 4.2& 2.8& 33.0 & 75.6 & 60.0 \\
    8-2 & 16.3 &  12.8& 7.0& 8.2& 8.0 & 33.0 & 25.0 \\
    9-1 & 97.0 &  38.0& 28.5& 100.0& 73.3 & 54.0 & 52.5 \\
    9-2 & 84.9 &  4.2& 10.0& 17.5& 73.3 & 63.3 & 53.6 \\
    9-3 & 123.1 &  0.0& 0.0& 1.7& 53.3 & 77.0 & 64.4 \\
    10-1 & 130.1 &  1.3& 24.5& 1.3& 17.0 & 76.0 & 30.76 \\
    10-2 & 132.1 &  0.3& 11.7& 6.0& 17.0 & 51.1 & 30.76 \\
    \hline
    Average & 49.26 &  25.22& 19.76& 23.40& 49.22 & 62.22 & 65.43  \\
    \hline
  \end{tabular}
  \label{app:table_diff_methods}
\end{table}

\subsection{Task Scores Across Various Model Sizes}

In \Cref{app:table_diff_sizes}, the average scores for models of different sizes are presented. In this experiment, the \textit{base} model was employed for the action generator, while the size of the sub-goal generator was varied across small, base, large, and x-large. The first four columns of the table display their respective scores. Additionally, using the \textit{small} model for the action generator, the experiment was repeated with various sizes of the sub-goal generator, and the results are shown in the second set of four columns. The last column illustrates the performance when both the action generator and the sub-goal generator are x-large, achieving the highest score.

\begin{table}
    \centering
    \caption{ 
The table displays the average scores for the models of different sizes. The first four columns depict scores when a model with a base size is used as the action generator. The subsequent four columns illustrate scores when the action generator is small size. The last column shows the results when both the action generator and sub-goal generator are x-large.}
    \begin{tabular}{|c|cccc|cccc|c|}
    \hline
 Action generator& \multicolumn{4}{c|}{Base} & \multicolumn{4}{c|}{Small} & X-Large\\ 
 \hline
\backslashbox[33mm]{Tasks}{Sg generator}&   Small&  Base&  Large&  X-Large& Small& Base& Large&X-Large &X-Large\\
  \hline       
 1-1&  0.2&  0.0&  0.0&  4.2 & 0.0& 0.0& 0.0&0.0 &16.55\\ 
 1-2&  0.0&  0.0&  0.0&  0.22 & 0.0& 0.0& 0.44&0.22 &17.11\\ 
 1-3&  0.0&  0.0&  0.0&  0.0 & 0.0& 0.0& 0.0&0.0 &0.0\\ 
 1-4&  0.1&  0.0&  0.0&  7.77 & 0.33& 0.0& 7.77&19.44 &20.44\\ 
 2-1&  10.0&  20.0&  100&  90.0 & 30.0& 10.0& 80.0&70.0 &100\\ 
 2-2&  13.2&  20.0&  20.0&  77.5 & 0.0& 20.0& 20.0&65.7 &77.5\\ 
 2-3&  0.0&  20.0&  0.0&  64.0 & 0.0& 20.0& 20.0&60.0 &45.0\\ 
 3-1&  0.0&  78.0&  84.0&  79.4 & 0.0& 48.0& 77.8&71.2 &100\\ 
 3-2&  10.6&  36.4&  47.0&  77.2 & 0.0& 27.8& 49.0&56.4 &77.2\\ 
 3-3& 26.0& 20.0& 22.0& 6.4 & 7.3& 10.5& 3.5&1.5 &28.0\\ 
 3-4& 40.0& 30.0& 40.0& 30.0 & 50.31& 30.0& 60.0&30.0 &47.05\\ 
 4-1& 100& 100& 100& 100 & 100& 100& 100&90 &100\\ 
 4-2& 100& 100& 100& 100 & 100& 100& 100&100 &100\\ 
 4-3& 100& 100& 92.5& 93 & 100& 100& 92.5&94.0 &100\\ 
 4-4& 100& 100& 100& 100 & 90.9& 100& 100&100 &100\\ 
 5-1& 66.2& 56.18& 100& 72.7 & 82.8& 67.1& 100&72.7 &100\\ 
 5-2& 31.8& 7.45& 39.6& 69.4 & 35.8& 0.0& 47.1&45.7 &70.0\\ 
 6-1& 3.12& 31.25& 43.62& 36.62 & 12.5& 28.12& 38.75&34.62 &56.25\\ 
 6-2& 20.0& 10.0& 66.66& 74.44 & 18.88& 32.22& 33.33&35.55 &91.11\\ 
 6-3& 23.0& 13.33& 33.66& 52.33 & 1.1& 6.33& 18.22&17.88 &54.88\\ 
 7-1& 80.0& 100& 100& 100 & 90.9& 100& 100&100 &100\\ 
 7-2& 20.0& 78.12& 80.0& 100 & 40.0& 50.0& 50.0&60.0 &100\\ 
 7-3& 100& 100& 100& 100 & 70.0& 70.0& 70.0&70.0 &100\\ 
 8-1& 11.6& 18.4& 60.0& 40.0 & 0.0& 18.4& 60.0&40.0 &60.0\\ 
 8-2& 0.0& 0& 0.0& 42.5 & 0.0& 0.0& 0.0&25.0 &67.5\\ 
 9-1& 28.0& 35.65& 24& 70 & 42.0& 44.0& 33.0&74.0 &84.0\\ 
 9-2& 26.0& 32.6& 23& 65 & 35.6& 44.0& 32.0&70.0 &82.0\\ 
 9-3& 27.0& 29.89& 29& 66 & 30.0& 43.0& 31.0&69.0 &83.0\\ 
 10-1& 27.0& 26.9& 30.1& 28.2 & 24.9& 25.2& 25.2&25.2 &28.2\\ 
 10-2& 30.2& 27.75& 30.4& 27.7 & 23.0& 19.5& 29.9&18.9 &30.1\\ 
 \hline
 Average& 33.13& 39.73& 48.85& 59.15 & 32.87& 37.13& 45.98&50.56 &67.86\\ 
\hline
\end{tabular}

\label{app:table_diff_sizes}
\end{table}

\subsection{Scores for Different Model Sizes Without sub-goals}
\Cref{app:table_diff_sizes_nosg} presents task scores without the utilization of sub-goals, akin to the Swift-only method but employing language models of varying sizes. All models are \textsc{flan-t5}.

\begin{table}[ht]
\centering
\begin{tabular}{|c|cccc|}
\hline
 Task Type&  Small&  Base&  Large&  X-Large\\
\hline
 1-1&  0.4&  0.0&  14.0&  14.0\\
 1-2&  11.8&  0.0&  22.88&  22.0\\
 1-3&  0.0&  0.0&  24.55&  20.0\\
 1-4&  7.6&  0.0&  0.88&  0.88\\
 2-1&  40.0&  80.0&  70.32&  91.66\\
 2-2&  43.2&  45.6&  6.56&  45.82\\
 2-3&  40.0&  43.0&  5.76&  41.0\\
 3-1&  59.8&  85.2&  85.2&  74\\
 3-2&  41.4&  51.0&  41.1&  43.4\\
 3-3& 19.2& 9.4& 64.88& 33.29\\
 3-4& 50.0& 33.0& 81.26& 58.75\\
 4-1& 100& 95.0& 100& 100\\
 4-2& 100& 97.7& 100& 100\\
 4-3& 80.0& 92.0& 80.0& 81.25\\
 4-4& 100& 95.0& 100& 100\\
 5-1& 64.2& 64.4& 24.0& 60.42\\
 5-2& 34.5& 35.0& 43.57& 42.53\\
 6-1& 12.75& 29.62& 26.0& 58.12\\
 6-2& 0.0& 11.11& 36.0& 34.11\\
 6-3& 3.6& 0.0& 11.33& 7.77\\
 7-1& 90.9& 100& 100& 100\\
 7-2& 40.0& 100& 100& 100\\
 7-3& 74.7& 78.34& 87.6& 100\\
 8-1& 8.2& 35.2& 46.0& 34.4\\
 8-2& 0.0& 0.0& 8.0& 41.0\\
 9-1& 20.0& 20.0& 40.7& 40.0\\
 9-2& 19.0& 20.0& 43.6& 43.0\\
 9-3& 20.0& 20.0& 44.52& 44.0\\
 10-1& 22.8& 23.0& 19.66& 23.0\\
 10-2& 22.9& 22.9& 20.22& 23.0\\
 \hline
 Average& 37.56& 42.88& 46.25& 52.88\\
 \hline
\end{tabular}
\caption{
Scores for each task without the utilization of sub-goals across various model sizes. All models are \textsc{flan-t5}.}
\label{app:table_diff_sizes_nosg}
\end{table}

\subsection{Results for \textsc{t5} Language Model With and Without sub-goals }

The results for the \textsc{t5} model, including \textsc{t5-large} and \textsc{t5-3b}, are presented in \Cref{app:table_t5}. The outcomes are shown for both scenarios—with and without the integration of sub-goals. In the experiments where sub-goals were employed, equivalent sizes were utilized for both the action generator and sub-goal generator.

\begin{table}[ht]
    \centering
    \begin{tabular}{|c|cc|cc|}
\hline
& \multicolumn{2}{c|}{No sub-goals} & \multicolumn{2}{c|}{With sub-goals}\\ 
 \hline
 Task Type&  Large&  X-large&  Large& X-large\\
 \hline
 1-1&  16.66&  0.2&  0.55& 18.0\\
 1-2&  0.0&  56.0&  5& 19.66\\
 1-3&  0.0&  8.55&  4.77& 0.0\\
 1-4&  0.0&  30.77&  16.22& 38.0\\
 2-1&  9.1&  91.66&  70.0& 70.0\\
 2-2&  37.8&  41.0&  20.0& 27.7\\
 2-3&  10.0&  30.0&  34.0& 44.0\\
 3-1&  85.2&  78.6&  39.2& 48.6\\
 3-2&  46.0&  55.2&  64.8& 30.6\\
 3-3& 50.2& 14.61& 1.0&31.76\\
 3-4& 30.0& 35.71& 40.0&50.0\\
 4-1& 100& 100& 100&100\\
 4-2& 100& 100& 100&100\\
 4-3& 90.0& 93.7& 100&100\\
 4-4& 100& 100& 100&74.33\\
 5-1& 100& 64.2& 91.5&100\\
 5-2& 8.9& 37.8& 2.1&100\\
 6-1& 41.62& 43.62& 21.87&31.25\\
 6-2& 15.55& 100& 55.55&73.33\\
 6-3& 4.77& 80.0& 29.2&35.0\\
 7-1& 90.0& 100& 100&100\\
 7-2& 100& 100& 30.0&100\\
 7-3& 80.0& 60.0& 100&100\\
 8-1& 55.4& 35.2& 60.0&69.0\\
 8-2& 0.0& 0.0& 0.0&0.0\\
 9-1& 20.0& 20.0& 36.0&33.33\\
 9-2& 21.0& 20.0& 34.2&36.1\\
 9-3& 22.0& 20.0& 37.1&39.0\\
 10-1& 27.9& 23.0& 29.69&30.4\\
 10-2& 28.0& 22.9& 28.75&28.0\\
 \hline
 Average& 43.00& 52.09& 45.05&54.26\\
 \hline
\end{tabular}
\caption{Scores for the \textsc{t5} model are depicted under two conditions: without sub-goals and with sub-goals.}
\label{app:table_t5}
\end{table}

\subsection{Results for random and semi-random sub-goals }
\label{app:rnd-sg}
We generated random sub-goals from the sub-goal space and used that instead of the sub-goal generator. The action generator remains the fine-tuned \textsc{Flan-t5-large}, as before. The results are displayed in \Cref{table:rand_sg_appendix}.

To further investigate the impact of the sub-goals, we conducted another experiment where we retained the sub-goals and solely modified their arguments, typically are objects or locations. Despite this adjustment, performance remained low, although slightly improved compared to random sub-goals. 

\begin{table}[ht]
    \centering
    \begin{tabular}{|c|c|c|} 
    \hline
 Task Type&  \parbox[t]{2cm}{Random\\ Sub-goal}& \parbox[t]{2cm}{Semi-random \\Sub-goal} \\ \hline
 1-1&  0& 0.5 \\ 
 1-2&  0& 1 \\ 
 1-3&  0& 2.22 \\ 
 1-4&  0& 0.6 \\ 
 2-1&  0& 3 \\ 
 2-2&  0& 3.1 \\ 
 2-3&  0& 2 \\ 
 3-1&  0& 2.8 \\ 
 3-2&  0& 4 \\ 
 3-3& 5.5&8 \\ 
 3-4& 0&1 \\ 
 4-1& 0&0 \\ 
 4-2& 58.3&81.4 \\ 
 4-3& 0&23.4 \\ 
 4-4& 0&2.5 \\ 
 5-1& 6.6&10.4 \\ 
 5-2& 21.9&60 \\ 
 6-1& 0&21 \\ 
 6-2& 15.55&23.33 \\ 
 6-3& 6.44&12 \\ 
 7-1& 10&60 \\ 
 7-2& 30&92 \\ 
 7-3& 0&0 \\ 
 8-1& 0&0 \\ 
 8-2& 0&0 \\ 
 9-1& 0&5.5 \\ 
 9-2& 0&5 \\ 
 9-3& 0&4 \\ 
 10-1& 0&14 \\ 
 10-2& 0&20 \\ \hline
 Average& 6.413&14.15 \\ \hline
    \end{tabular}
    \caption{Scores for the random and semi-random sub-goals with fine-tuned \textsc{Flan-t5-large} action generator.}
    \label{table:rand_sg_appendix}
\end{table}

\subsection{Sub-goal Retrieving}
\label{app:ret-sg}
Here are the scores representing the ability to retrieve sub-goals for the sub-goal generators. There are two sets of the experiments: 1- Random/semi-random first sub-goals, 2-Random/semi-random sub-goals during interactions. The scores for each task and experiment are presented in Table \ref{tab:ret_sg}.

\begin{table}
\centering
\begin{NiceTabular}{|c|>{\columncolor{red!10}}c|c|>{\columncolor{red!10}}c|c|}
\hline
 Task Type&  \parbox[t]{3cm}{Semi-random First\\ Sub-goal}&\parbox[t]{2.5cm}{Random First\\ Sub-goal}  & \parbox[t]{3.5cm}{Semi-random Sub-goal\\ at 10 steps}& \parbox[t]{3cm}{Random Sub-goal \\at 10 steps}\\   \hline
 1-1&  36.2&33 &  8.22& 8.8\\
 1-2&  19.33&20.11 &  31.44& 30.55\\
 1-3&  20.77&28.22 &  33& 25.4\\
 1-4&  73&26.44 &  26.88& 16\\
 2-1&  40&31.2 &  28.4& 48.4\\
 2-2&  16.4&17.2 &  12.2& 28.1\\
 2-3&  17&18 &  12& 28\\
 3-1&  78&32.8 &  78& 50.6\\
 3-2&  78.2&28.6 &  56.2& 62.4\\
 3-3& 6&6.5 & 13.4&10\\
 3-4& 12&50.5 & 0&36.8\\
 4-1& 100&70.8 & 92.5&90\\
 4-2& 100&82.5 & 100&97.3\\
 4-3& 80&57.5 & 85.5&81.5\\
 4-4& 100&79.8 & 97.4&60\\
 5-1& 100&51.7 & 53.1&33\\
 5-2& 41&23.8 & 31.8&5\\
 6-1& 31.65&42.5 & 26&24.25\\
 6-2& 75.55&77.77 & 54.44&71.11\\
 6-3& 52.88&44.77 & 32.88&17.6\\
 7-1& 100&80 & 100&70\\
 7-2& 100&90 & 100&80\\
 7-3& 80&70 & 70&50\\
 8-1& 0&20 & 20&0\\
 8-2& 25&0 & 0&0\\
 9-1& 52&15 & 27&31\\
 9-2& 50&14 & 26&29\\
 9-3& 51&16 & 25&28\\
 10-1& 27.5&20.9 & 29.5&1.8\\
 10-2& 29.6&24.6 & 29&13.6\\   \hline
 Average& 53.10&39.14 & 43.33&37.61\\   \hline
    \end{NiceTabular}
    \caption{Scores for retrieving sub-goals are displayed. The first two columns depict the random and semi-random generation of the first sub-goals only. The second pair of columns illustrates the scores obtained when sub-goals are generated randomly or semi-randomly every $10$ steps.}
    \label{tab:ret_sg}
\end{table}

\subsection{Generalization experiment}
In this experiment, we trained the model on a subset of tasks and evaluated it on test variations from all tasks. For each scientific topic, we selected one or two tasks for training and reserved the remaining tasks for evaluation. We compared our algorithm against the Swift-only baseline. The scores for each task are displayed in Table \ref{tab:held_out}.

\begin{table}
\centering
\begin{tabular}{|c|c|c|} 
\hline
 Task Type &  With sub-goal &  Swift-only \\ \hline \hline
 \hl{1-1}&  6&  5\\ 
 1-2&  1.3&  0\\ 
 \hl{1-3}&  2.2&  0\\ 
 1-4&  9.33&  74.33\\ 
 \hl{2-1}&  38.9&  100\\ 
 \hl{2-2}&  20&  33.9\\ 
 2-3&  20&  0\\ 
 \hl{3-1}&  35.4&  100\\ 
 3-2&  14.6&  0\\ 
 \hl{3-3}& 16.9& 19.6\\ 
 3-4& 20& 15.5\\ 
 \hl{4-1}& 100& 100\\ 
 \hl{4-2}& 95& 100\\ 
 4-3& 39.2& 0\\ 
 4-4& 65& 53\\ 
 \hl{5-1}& 90.7& 81.8\\ 
 5-2& 50& 2\\ 
 \hl{6-1}& 69.75& 23.75\\ 
 6-2& 62.22& 30\\ 
 \hl{6-3}& 38.77& 2.5\\ 
 \hl{7-1}& 100& 100\\ 
 \hl{7-2}& 100& 100\\ 
 7-3& 10& 16.6\\ 
 \hl{8-1}& 40& 64.4\\ 
 8-2& 0& 0\\ 
 \hl{9-1}& 44.5& 20\\ 
 \hl{9-2}& 40& 19\\ 
 9-3& 41& 5\\ 
 10-1& 20.6& 28.6\\ 
 \hl{10-2}& 28.7& 1.9\\ \hline
 Average score& 40.63& 36.56\\ \hline
 Average score seen tasks& 50.51& 52.85\\\hline
 Average score unseen tasks& 27.72& 15.25\\\hline
    \end{tabular}
    \caption{The table displays the scores for the generalization experiment. The highlighted tasks in yellow called ``seen tasks'' which are the ones selected for the training.
    }
    \label{tab:held_out}
\end{table}

\end{document}